%% file: main.tex
\documentclass[10pt,twocolumn,letterpaper]{article}
\pdfoutput=1    
 
\usepackage{iccv}
\usepackage{times}
\usepackage{epsfig}
\usepackage[mode=buildnew]{standalone}
\usepackage{graphicx}

\usepackage{amsmath}
\usepackage{amssymb}
\usepackage{url}

\usepackage{subcaption}
\usepackage[accsupp]{axessibility}  
\usepackage[pagebackref=true,breaklinks=true,letterpaper=true,colorlinks,bookmarks=false]{hyperref}

\usepackage[breaklinks=true,bookmarks=false]{hyperref}

\iccvfinalcopy 

\def\iccvPaperID{16} 

\ificcvfinal\fi

\begin{document}

\title{Description of Corner Cases in Automated Driving: Goals and Challenges}

\author{Daniel Bogdoll$^{1}$\textsuperscript{\textasteriskcentered} \and
        Jasmin Breitenstein$^{2}$\textsuperscript{\textasteriskcentered} \and
        Florian Heidecker$^{3}$\textsuperscript{\textasteriskcentered} \and
        Maarten Bieshaar$^{3}$ \and
        Bernhard Sick$^{3}$ \and
        Tim Fingscheidt$^{2}$ \and
        J. Marius Zöllner$^{1}$
}

\maketitle
\footnotetext[1]{Daniel Bogdoll and J. Marius Zöllner are with Technical Cognitive Systems, FZI Research Center for Information Technology, Schönfeldstraße 8, 76131 Karlsruhe, Germany, {\tt\small \{bogdoll, zoellner\}@fzi.de}}
\footnotetext[2]{Jasmin Breitenstein and Tim Fingscheidt are with Institute for Communications Technology, Technische Universität Braunschweig, Schleinitzstraße 22, 38106 Braunschweig, Germany, {\tt\small \{j.breitenstein, t.fingscheidt\}@tu-bs.de}}
\footnotetext[3]{Florian Heidecker, Maarten Bieshaar and Bernhard Sick are with Intelligent Embedded Systems, University of Kassel, Wilhelmshöher Allee 73, 34121 Kassel, Germany, {\tt\small \{florian.heidecker, mbieshaar, bsick\}@uni-kassel.de}}
\begingroup\renewcommand\thefootnote{\textasteriskcentered}
\footnotetext{These authors contributed equally}
\endgroup

\ificcvfinal\thispagestyle{empty}\fi

\input{content}

\section*{Acknowledgment}
This work results from the project KI Data Tooling (19A20001O, 19A20001J, 19A20001M) funded by the German Federal Ministry for Economic Affairs and Energy (BMWI).

{\small
\bibliographystyle{ieee_fullname}
\bibliography{bibtex}
}

\end{document}

%% file: content.tex
\begin{abstract}
Scaling the distribution of automated vehicles requires handling various unexpected and possibly dangerous situations, termed corner cases (CC). Since many modules of automated driving systems are based on machine learning (ML), CC are an essential part of the data for their development. However, there is only a limited amount of CC data in large-scale data collections, which makes them challenging in the context of ML. With a better understanding of CC, offline applications, e.g., dataset analysis, and online methods, e.g., improved performance of automated driving systems, can be improved. While there are knowledge-based descriptions and taxonomies for CC, there is little research on machine-interpretable descriptions. In this extended abstract, we will give a brief overview of the challenges and goals of such a description.

\end{abstract}


\section{Introduction}\label{sec:Intro}
Corner cases (CC) are data that occur infrequently or represent a critical situation and are only available in datasets to a limited extent, if at all. However, for machine learning (ML), CC are important as they are required for training, verification, and improved performance of ML models during inference within automated driving systems. 

\begin{figure}[t]
    \centering
    \includegraphics[width=0.48\textwidth]{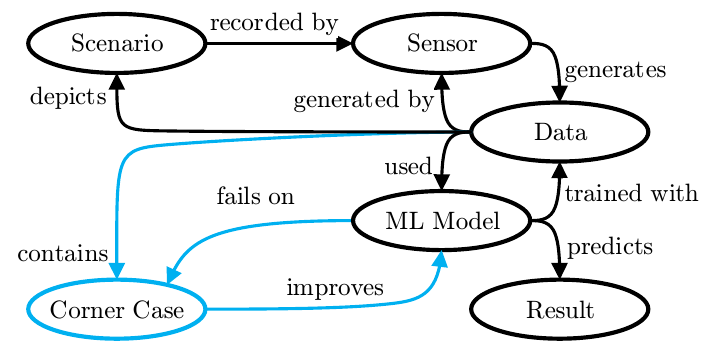}
    \caption{ML pipeline in automated driving with an emphasis on corner cases and their relation to data and ML models based on corner case descriptions.}\label{fig:cc_in_ml}
\end{figure}

While automated vehicles will always encounter new CC, it is crucial to have training data with a wide variety of scenarios \cite{karpathy2021CVPR2021Workshop}, including CC. This way, failures and unexpected behavior of ML models during inference can be reduced. Such failures can occur on several levels of the system, such as missing detections or incorrect behavior predictions \cite{Heidecker2021b}. To achieve a good \emph{online} performance of automated driving systems, it is necessary also to consider the previous \emph{offline} development steps, such as dataset engineering. Since CC play a crucial role in both fields, corner case descriptions (CCD) act as a basis.

CC can be described either by experts or directly by the limits of a method \cite{Heidecker2021b}. On one hand, they can be divided into the \emph{sensor layer}, \emph{content layer} and \emph{temporal layer}. \emph{Sensor-layer} CC are directly connected to a sensor, e.g., dead pixels. The \emph{content layer} ranges from domain shifts and object-related CC to events on the scene\footnote{We follow the definitions of scene and scenario by Ulbrich~et~al. \cite{Ulbrich:2015d}, where a scene is a snapshot, and a scenario consists of successive scenes} level. Finally, the \emph{temporal layer} includes all scenario-related CC.

On the other hand, \emph{method-layer} CC occur when an ML model reaches its limits, as when failures occur due to insufficient model knowledge caused by incomplete training data \cite{Heidecker2021b}. Even models as DeepXplore \cite{Pei2017}, which automatically create CC, are unable to create descriptions of those.

Describing CC with a proper description language and using those descriptions are fields with many challenges and applications. CCD are necessary for different aspects of the \emph{offline} development pipeline in automated driving and the subsequent \emph{online} validation and deployment tasks.

A major research question is how to provide a valid description of CC in co-existence with established scenario description languages. This can be useful in \emph{offline} applications to develop automated driving systems:
First, existing data can be automatically described based on a CCD, including the coverage of datasets in respect to CC. Valid descriptions of CC are necessary to generate synthetic data or to record real data. Subsequently, CCD can be utilized for \emph{online} applications. In the validation of automated driving systems, CCD can simulate variations of certain CC to cover as many variants of a situation as possible. During the deployment of automated vehicles, perception and fusion can benefit from such descriptions to determine which sensors are best equipped to detect certain CC, or even serve as an interface to the prediction or planning tasks. 

An insight into how we envision CC within a description language is shown in Figure \ref{fig:cc_in_ml}. Scenarios, either conceptual or from the real world, are embedded within the data recorded by sensors. Only here can CC be detected. An ML model to perform perception tasks might fail online on CC data, but can be improved, if CC can be detected and properly understood. Therefore, CCD have the potential to increase the performance of ML models.

Such a CCD comprises various aspects, some of which are already contained in established scenario description languages \cite{ASAM2021, PEGASUSProject2019}. However, there are significant research gaps regarding CCD which we discuss in the following.

Our contributions of this extended abstract are: We provide an overview over the state of the art of multiple research areas related to CC and CCD. We formulate six research questions to outline topics for future work, and describe the potential of CCD based on three use cases in regard to our research questions.
 
The remainder of this article is structured as follows: Section~\ref{sec:RelatedWork} provides an overview of related research.  Section~\ref{sec:RQ} formulates our research questions. Section~\ref{sec:use_cases} includes use cases. Finally, Section~\ref{sec:conclusion} provides a summary and an outlook.

\section{Related Work}\label{sec:RelatedWork}
So far, CCD have received little attention in research. However, they have many connections to other research areas, such as scenario descriptions, which are used in validation processes to generate suitable validation scenarios, and are currently limited in their ability to describe CC. In addition, there are related research areas such as CC detection, generation, and dataset engineering that benefit from established CCD.

\subsection{Scenario Description}\label{sec:SenarioDescription}
In the field of automated driving systems, all real world data is embedded into scenarios, even though CC can also occur and be detected on lower levels \cite{Heidecker2021b}. Scenarios can be described on different levels of abstraction, namely \emph{functional, logical and concrete} \cite{menzelScenariosDevelopmentTest2018}. In the following, we will give an overview of scenario descriptions as we aim to combine existing scenario descriptions with a CCD. Since every CC stems from a scenario, as shown in Figure \ref{fig:cc_in_ml}, scenario description is the field with the closest relation to CCD. Here, the basic entities in a scene are described by concrete scenario description languages. A CCD expands on this and describes the CC, based on the categories of \cite{Heidecker2021b}. One example might be to classify a scene or scenario by a CCD with a level, a type, and a machine-interpretable description, including relevant entities.

\subsubsection{PEGASUS Layer Model}
The PEGASUS model \cite{PEGASUSProject2019,Weber2019}, a generic description model for automated driving scenarios, contains six layers, ranging from road-level to digital information.
The first layer deals with road geometry and its characteristics, while the second layer includes traffic infrastructure. Temporary modifications are considered in the third layer, such as movable road infrastructure. All other movable objects, vehicles, and vulnerable road users (VRU) are described in the fourth layer. Environmental influences, e.g., weather, are summarized in the fifth layer. Finally, the sixth layer deals with digital information and describes V2X (Vehicle-to-everything) information and digital maps. The language aims to describe scenarios on all three abstraction levels \cite{menzelScenariosDevelopmentTest2018}.

\subsubsection{Open Scenario \& Open Drive}
The key feature of OpenScenario is to describe traffic scenarios consisting of multiple actors, focusing on concrete scenarios. Thereby, the temporal succession of actions such as an overtaking maneuver is clearly described. A standard for the scenario description is presented in ASAM OpenScenario \cite{ASAM2021}. However, OpenScenario does not provide a driver model or advanced motion dynamics for vehicles. ASAM OpenDrive \cite{ASAM2021a} is a standard and describes the road network, which contains the road geometry, lanes, and signals. The surface profile of the road is standardized within ASAM OpenCRG (Curved Regular Grid) \cite{ASAM2021b}. Combining these components, automated in-the-loop vehicle simulation of dynamic and static content is possible.

\subsubsection{Ontology-Based Descriptions}
Ontologies have the ability to inherit structure and terminology from taxonomies to describe relations, attributes, and general concepts of and between classes. As shown in \cite{asmarAWAREOntologySituational2020}, their representations can vary from meta-levels to very detailed, even perception-related attributes and therefore seem suited for all levels. As demonstrated in \cite{Kuhnt2020_1000118076}, they can be used to describe traffic scenes, including temporal context. This way, dynamic objects and details about their configurations, intentions, and information about the road layout can be combined for a holistic scenario description. Once an ontology is designed, instances can be created to describe concrete situations.

\subsubsection{Operational Design Domain}
An operational design domain (ODD) describes the operating domain of an automated driving system, that has been designed to limit the situations that possibly occur. This can act as a baseline to take into account combinations of possible operating conditions during validation \cite{Koopman2019}. Morphological boxes, also called \textit{zwicky boxes}, provide a possibility to describe an ODD using possible dimensions, i.e. relevant features, and corresponding attributes, i.e. value ranges \cite{Bitzer2020,Gladisch2020}. Describing CC in an ODD is difficult as many per definition would not be included in the description in the first place. In the case of \textit{zwicky boxes}, combinatorial anomalies can be created, but, e.g., any CC based on temporal context is out of range for the description.

\subsection{Corner Case Detection and Generation}
The safety of automated driving functions depends highly  on reliable CC detection both for curating suitable training and test datasets and as an online alert system.

\paragraph{Corner Case Detection:}
Depending on conceptual levels \cite{Heidecker2021b, Breitenstein2020}, different CC exists and, hence, implicating different detection methods. Certain types of CC can be detected from single image frames, such as physical-level CC, e.g., glare situations \cite{Jatzkowski2018,Dhananjaya2021}, or content-layer CC, e.g., unknown objects in the traffic situation \cite{xia2020}. Temporal-layer CC encompass an entire time span. Hence, detection methods need to take temporal data into account, such as video sequences \cite{Bolte2019b}. Such detection systems typically provide metric-based assessments of situations, such as shown in \cite{Breitenstein2020}. Such metrics, without a context, typically do not contribute to an improved understanding of the situation. A CCD is a potent tool to embed such metrics in a context.

\paragraph{Corner Case Generation:}
CC generation strategies can be divided into the classes \emph{model-based, data-driven and scenario-based} \cite{tuncaliRequirementsDrivenTestGeneration2020a}. Tuncali~et~al. \cite{tuncaliRequirementsDrivenTestGeneration2020a} propose a model-based generation method that is able to generate CC based on a combination of covering arrays and requirement falsification methods. Chou et al. \cite{chou2018UsingControlSynthesis} propose a similar system, which determines corner case situations for an automated vehicle which are critical, but not impossible to handle. Data-driven methods, such as shown in \cite{mollerOutofdistributionDetectionGeneration2021}, rely on existing datasets and samples from the available data. M\"oller~et~al.~\cite{mollerOutofdistributionDetectionGeneration2021} generate potential CC by learning a latent space based on available data and subsequently generate out-of-domain scenarios, which are still close to the actual in-distribution samples and therefore likely to be realistic. Finally, scenario-based methods mostly utilize descriptions of experts or compiled within accident reports. Pretschner~et~al. \cite{pretschnerTestsFuerAutomatisierte2021a} propose a \emph{simulation-based heuristic search} to generate similar scenarios as shown in \cite{chou2018UsingControlSynthesis}, which are basically critical, but do not necessarily lead to a collision. \cite{klueckUsingOntologiesTest2018} shows a methodology, which samples CC based on an ontology, which describes the environment of a vehicle. CC generation is dependent on sufficient and precise CCD to generate suitable data.

\subsection{Dataset Engineering}

As dataset engineering, we identify the task of putting together a suitable dataset for the underlying perception function. When recording new data, CCD can either act as a trigger a recording during fleet operation \cite{karpathy2021CVPR2021Workshop} or as a formal description for controlled records.
When considering existing datasets, the challenge is to describe them, quantify their coverage, and measure their diversity. Sadat~et~al. \cite{Sadat2021} define various complexity measures for dataset diversity, taking into account the relevance of certain traffic situations. In \emph{active learning}, one selects interesting data for additional labeling to keep costs of labeling low. As an example of using CC in this research area, Dhananjaya~et~al. \cite{Dhananjaya2021} applied active learning for weather-related CC. In contrast to general dataset coverage, such active learning methods are based on the underlying model \cite{Sadat2021}. CCD can aid the task of dataset engineering as they give a way to describe the CC coverage of a dataset and to filter for certain CC. This then allows determining the CC coverage of ML models based on their training dataset.

\section{Research Questions}\label{sec:RQ}
The concept of CCD has to overcome various challenges before it can leverage its full potential. We summarize these points with the following research questions (RQ), focusing on the description itself, followed by \emph{offline} and subsequent \emph{online} applications.


\textbf{Description (\emph{RQ1}):} \textit{How to describe CC on different levels of abstraction?} -- 
Heidecker~et~al. \cite{Heidecker2021b} present a high-level categorization of CC in different layers and levels. We aim to combine this with a concrete scenario description language, as outlined in Section \ref{sec:SenarioDescription}, or meta-data to build a complete description of CC. This makes it possible for us to use existing elements of scenario description languages, which describe the scene in its basic elements. We believe an add-on should expand on this and address the CC, whereby we differentiate between the categories of \cite{Heidecker2021b} in the description, and then specifying the details. To improve and complete the description languages, data-driven approaches should be utilized, e.g. adding missing object classes. Based on a CCD, we see multiple \emph{offline} applications, which we address with the following RQ:\\

\textbf{Automation (\emph{RQ2}):} \textit{How to automatically describe data that contain CC?} --
For an efficient generation of scenarios containing CC, and other tasks, it is important to not only have a valid CCD but to be able to extract them automatically. Methodologies to detect CC on sensor and object data levels are shown in \cite{Breitenstein2020,toettel2021reliving}. Having a machine-readable CCD as an interface will allow transferring algorithmic outputs into a CCD automatically. Thus, we can analyze existing datasets and enhance them with CCD for every single frame. While there is early research in this field, no extensive methodology has been developed to describe CC precisely. Combining existing CC levels \cite{Heidecker2021b} with efficient scenario description languages such as \textit{zwicky boxes}, we aim to provide an automatic way to describe CC.

\textbf{Data Generation (\emph{RQ3}):} \textit{How to generate or record CC from descriptions?} --
CCD will be used to build up scenario catalogs as concrete guidelines for human-performed data recordings or as input for data augmentation tools, such as \cite{2018-01-1080}, or simulation environments, e.g., CARLA \cite{dosovitskiy2017carla}. They are also potent trigger descriptions for fleet recordings \cite{karpathy2021CVPR2021Workshop}. Feeding automatically generated CCD from RQ2 into a simulation tool, enables automatic description and generation of corner cases.

\textbf{Dataset Analysis (\emph{RQ4}):} \textit{How to determine the coverage of CC in training data?} --
We aim to describe the coverage of a CC detector or ML model. If we describe the training/testing data with a CCD (e.g., by including the context), this will represent the content and thus the cases covered in the dataset. Transferring this description to the ML model trained and tested on the dataset, we can obtain the coverage that the ML model has or should have via evaluating its performance achieved on that specific dataset and coverage.
Conversely, we also know what it does not cover. Additional training data would now be selected to achieve better generalization and coverage of the ML model. In case some CC are missing to fulfill the required coverage, the CCD is used to explicitly record or generate these CC scenarios synthetically, see \emph{RQ3}. These \emph{offline} development stages enable the \emph{online} deployment of automated vehicles in the real world, which we address with the following research questions:\\

\textbf{Automated Driving (\emph{RQ5}):} \textit{How to utilize CCD for driving automation?} --
CCD can be utilized to provide proper situation understanding for perception tasks, such as sensor fusion, since not all sensors are equally well suited to detect CC. CCD as a machine-readable format is an ideal candidate as an interface between the perception and later stages of the automated driving pipeline, transferring CC-related information to models such as prediction or planning. One example is to describe CC based on relevant objects instead of a static metric, which is a more actionable interface. While not much research regarding this topic can be found in academia, in industry, the activation of remote assistance is often triggered as soon as CC are detected or the ODD is exited \cite{ackermanWhatFullAutonomy2021}.

\textbf{Testing (\emph{RQ6}):} \textit{How to validate and verify ML systems based on CCD?} --
Scenario description languages are currently also used for automated testing and verification while not providing sufficient information about the coverage of CC within these scenarios. With CCD as an extension, this will become possible, improving upon existing approaches. 

\section{Addressed Use Cases}\label{sec:use_cases}

The CCD and associated RQ show why a description language and the corresponding structuring of CC are necessary. To demonstrate the potential of a CCD, we build upon the RQ and show their application-related relevance in the real world.

\begin{figure}[h]
    \centering
    \includegraphics[width=0.5\textwidth]{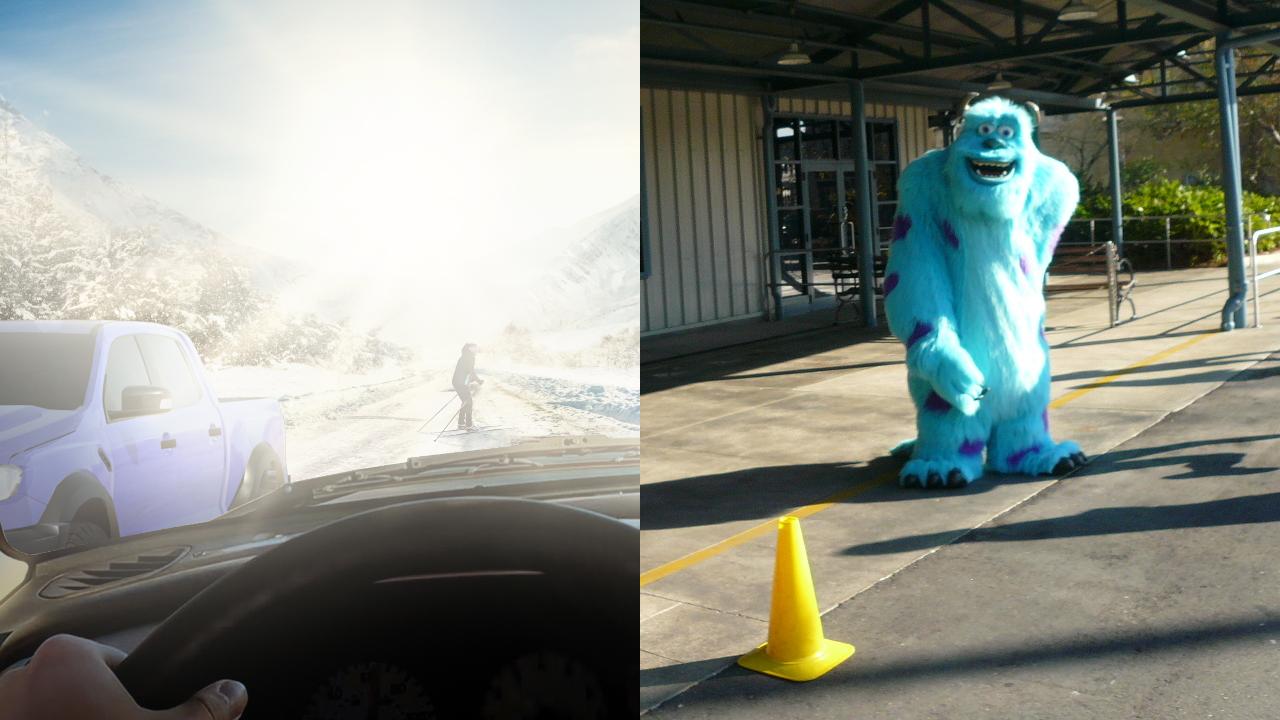}
    \caption{A visually challenging scene due to glare \cite{Heidecker2021b} (l) and an unknown object in the field of view \cite{chan2021segmentmeifyoucan} (r).}
    \label{fig:glare_unknown}
\end{figure}

For this reason, we present the following three use cases and show how they would benefit from the CCD. We sketch out conceptual CCD for the use cases, focused on VRU.

\begin{figure*}[t]
    \centering
    \begin{subfigure}[b]{0.3\textwidth}
        \centering
        \includegraphics[width=\textwidth]{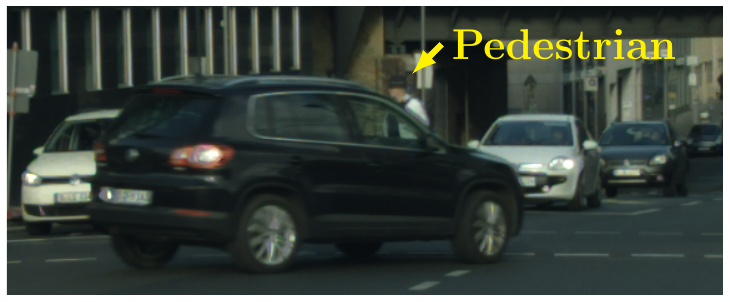}
        \label{subfig:04}
    \end{subfigure}
    \begin{subfigure}[b]{0.3\textwidth}
        \centering
        \includegraphics[width=\textwidth]{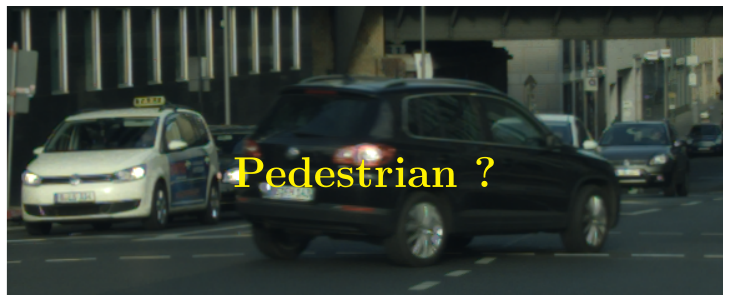}
        \label{subfig:08}
    \end{subfigure}
    \begin{subfigure}[b]{0.3\textwidth}
        \centering
        \includegraphics[width=\textwidth]{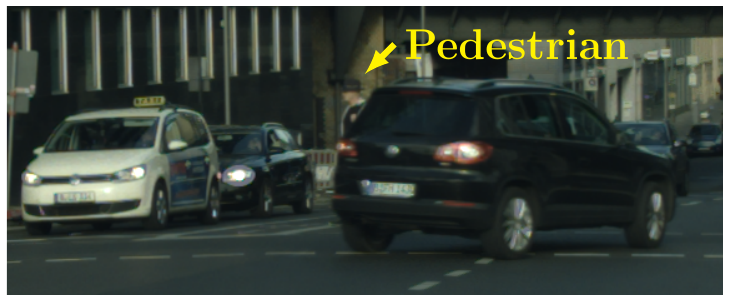}
        \label{subfig:11}
    \end{subfigure}
    \caption{Sequence of temporarily occluded road users \cite{Cordts2016}.}
    \label{fig:occluded_object}
\end{figure*}

\subsection{Glare Effect}

Being blinded by the sun, an artificial light source, or even reflected light, hence not seeing the actual object (e.g., pedestrian), is a situation that everyone is familiar with (see Figure \ref{fig:glare_unknown} (l)). ML models that work with image data also have problems with being blind, as it harms the object detection or prediction of any kind. The glare can lead to a failure of the driving function and, in the worst case, to an accident. First, a CCD is necessary to allow tailored handling of the situation. A CCD including a scenario description, as in \emph{RQ1}, of Figure \ref{fig:glare_unknown} (l) would conceptually be: CC-level - physical level, CC-type - global outlier, scene - winter, sun - low (over the road), visibility - poor, vehicles - yes, vehicle1: - on the other lane, coming in our direction, VRU - yes, VRU1: type - pedestrian, located - ahead on the road, moving - from left to right. The combinatorics of an extensive CCD allow generating new scenarios with glare effects \emph{RQ3}. Regarding \emph{RQ5}, based on the CCD, the sensor fusion layer could put more emphasis on other sensors, such as LiDAR or RADAR, as the system has gained an awareness that the image data has currently reduced information richness. A CCD can also be used to describe test cases that an ML model must pass for the validation \emph{RQ6}.

\subsection{Unknown Objects}

In road traffic, an unmanageable number of scenarios occur every day. Some of these scenarios contain objects (see Figure \ref{fig:glare_unknown} (r)), which were never or only to a small extent included in the training data of an ML method. In addition, new objects are constantly being created and modified, which negatively affects object detection reliability and leads to an increased risk of accidents when used in autonomous driving functions. According to \cite{Heidecker2021b}, this group can be described as object-level CC and is included in the content layer. The challenge is to identify these CC and make them manageable for the driving function. A precise CCD \emph{RQ1} can have a positive influence on object detection in several ways. A conceptual description of the CC in Figure \ref{fig:glare_unknown} (r) is: CC-level - object level, CC-type - single-point anomaly, VRU - yes, VRU1: type - pedestrian, located - next to the road, moving - no, clothing - costumed, direction of view - towards us. 

Blum~et~al. \cite{blumFishyscapesBenchmarkSafe2019} demonstrate, how \emph{RQ3} can be addressed with automated CC generation for unknown objects, resulting in similar data as shown in Figure \ref{fig:glare_unknown} (r). 
With \emph{RQ4}, based on a CCD, the typically broad category of unknown objects can be improved drastically, leading to a much better understanding of such objects. This would be convenient for object detection since we can quantify the ML model's coverage via the description. Also, CC are important for unknown objects because they are needed to validate the ML Model (\emph{RQ6}).

\subsection{Occluded Objects}
Situations concerning temporarily occluded VRU are often accompanied by an increased risk for collision due to the uncertainty about the VRU reappearance. As such a situation covers an entire period, it describes a CC on the temporal layer \cite{Heidecker2021b}. A conceptual description of the presented scenario in Figure \ref{fig:occluded_object} would be: CC-level - scenario level, CC-type - risky scenario, infrastructure - intersection, vehicles - yes, vehicle1: - coming from left and left turn, VRU - yes, VRU1: type - pedestrian, located - opposite street side, moving - crossing from right to left, visible - true. This CCD applies to the left and right images in Figure \ref{fig:occluded_object}. For the middle image, the scene description would not contain a pedestrian even though we know one is present because temporal context is missing. The CCD now changes to visible - false but keeps the pedestrian as an entity.

The challenge is to ensure no critical situations appear, which could not be handled by the automated driving system. Based on an automatically created CCD (\emph{RQ1}, \emph{RQ2}), the driving function (\emph{RQ5}) would be aware of the occluded VRU. A CCD also allows the automatic generation of synthetic scenarios with occluded VRUs (\emph{RQ3}). 

\section{Conclusion and Outlook}\label{sec:conclusion}
In this work, we formulated research questions for the application and need for valid corner case descriptions (CCD) focusing on machine learning (ML) models. In this regard, we reviewed existing scenario description languages based on their ability to describe corner cases (CC) and related research fields, such as CC generation. Based on our formulated research questions, we have shown the potential of a CCD based on three challenging situations within the domain of automated driving in the real world. We propose a combination of high-level knowledge-based CCD, low-level scenario-based CCD, and metric-based CC severity assessments as the basis for a machine-interpretable CCD, which will be the focus of future research.